\documentclass[11pt]{article}
\usepackage[a4paper]{geometry}
\usepackage{clic2017}
\usepackage{times}
\usepackage{url}
\usepackage{latexsym}
\usepackage{graphicx}	
\usepackage{xcolor}
\usepackage{calrsfs}
\DeclareMathAlphabet{\pazocal}{OMS}{zplm}{m}{n}

\newcommand{\Lb}{\pazocal{L}}
\usepackage[tracking=false]{microtype}

\title{Detecting Early Onset of Depression from Social Media Text using Learned Confidence Scores}

\author{Ana-Maria Bucur\\
  University of Bucharest, Romania \\
  {\tt ana-maria.bucur@drd.unibuc.ro} \\\And
  Liviu P. Dinu \\
  University of Bucharest, Romania \\
  {\tt ldinu@fmi.unibuc.ro} \\}

\date{}

\begin{document}
\maketitle
\begin{abstract}
  \textbf{English.}  Computational research on mental health disorders from written texts covers an interdisciplinary area between natural language processing and psychology. A crucial aspect of this problem is prevention and early diagnosis, as suicide resulted  from  depression being the second leading cause of death for young adults. In this work, we focus on methods for detecting the early onset of depression from social media texts, in particular from Reddit. To that end, we explore the eRisk 2018 dataset and achieve good results with regard to the state of the art by leveraging topic analysis and learned confidence scores to guide the decision process. \footnote{Copyright \copyright2020 for this paper by its authors. Use permitted under Creative Commons License Attribution 4.0 International (CC BY 4.0).}
%   Our analysis paves way to more in depth exploration of detection of mental illnesses from social media interactions.
\end{abstract}

\section{Introduction}

Mental illnesses are a common problem of our modern world. More than one in ten people was living with mental health disorders in 2017 \cite{ritchiemental}, with women being the most affected. These disorders affect people's way of thinking, mood, emotions, behaviour and their relationships with others. Most mental illnesses remain undiagnosed because of the social stigma around them.

Depression is one of the main causes of disability globally \footnote{\url{https://www.who.int}}, it affects people of all ages. Prevention is used to reduce depression and to save the lives of people at risk of suicide, but prevention is only limited to raising awareness and programs to cultivate positive thinking in case of depression and monitoring people who attempted suicide or self-harm.

With the rise in social media use, more computational efforts are made to detect mental illnesses such as depression \cite{de2013predicting} and PTSD \cite{coppersmith2015clpsych}, but also to detect misogyny \cite{anzovino2018automatic}, irony and sarcasm \cite{khokhlova2016distinguishing} from users' texts.

People tend to talk more about their emotions and mental health problems online and to seek support. The sources of mental health cues used for detection are Twitter, Facebook, Reddit and forums \cite{calvo2017natural}. Reddit\footnote{\url{https://www.reddit.com/}} is a social media site very similar to forums. It is organized in subreddits with specific topics, some dedicated to mental health problems. The use of throwaway accounts to maintain anonymity promotes disclosure, and users are more likely to share problems they have not discussed with anyone before. The use of these accounts makes it difficult for users to receive more social support because the majority of them are used only for one post \cite{calvo2017natural}.

In this work, we choose to tackle the problem of detecting early onset of depression from users' posts on social media, specifically from Reddit. As such, we explore the eRisk 2018 dataset through topic analysis by means of Latent Semantic Indexing \cite{lsa-paper} and learned out-of-distribution confidence scores \cite{devries2018learning}. Due to the nature of the dataset, we repurpose the learned confidence score to make a decision on whether to label the user as depressed or non-depressed or to wait for more data, as test chunks were progressively released every week.

\section{Related Work}

Recent studies for depression detection from text are reviewed by Guntuku et al. \cite{guntuku2017}. People diagnosed with mental illnesses from the datasets are identified using screening surveys, self-reported posts about diagnosis from social media or by their membership in different forums related to mental health. The most used features are topic modelling, n-grams, Linguistic Inquiry and Word Count (LIWC), emotion and metadata. The most used methods are Support Vector Machines (SVM), Logistic Regression, Random Forests and Neural Networks.

Coppersmith et al. \shortcite{coppersmith2016exploratory} show the differences in emoticons use between suicidal users and controls, neurotypicals using emojis with a much higher probability than a user before an attempt. Prior to the suicide attempt, the users at risk tend to use a more self-focused language, same as the people diagnosed with depression. The authors highlight different changes in post emotions before and after the suicide attempt: subjects have a higher incidence of anger and sadness posts prior to a suicide attempt, while fear and disgust tend to decrease after the attempt. People are also more likely to talk about suicide after an attempt than before it.

Sekuli{\'c} et al. \shortcite{sekulic2018not} indicate that users diagnosed with bipolar disorders use more first-person singular pronouns, same as depressed people. They also use more words associated with emotions; words associated with positive emotions as well as words associated with negative emotions explained by alternating episodes of mania and depression.

Nalabandian el al. \shortcite{nalabandian2019depressed} show that depressed persons tend to use more negative words and a self-focused language when writing about their interactions with a close romantic partner than when writing about other people around them. This is because people experience different symptoms of mental illness based on their interactions with other people.

Loveys et al. \cite{loveys2018cross} show the differences in language use of users with depression from different cultures to avoid cultural biases. Even if depression affects people all over the world, the way they experience and express it is shaped by their cultural context. The authors show that White and Black or African American people use more negative emotions, while Asian and Pacific Islander people tend to inhibit negative emotions. Hispanic or Latino people use both negative and positive emotions to a greater extent. Users from some ethnic groups does not address mental health issues as much as the others and this can make the depression task more difficult. After topic modeling, the topics are created for each cultural group. The words from each topic vary for each ethnic group, showing that people from each group discuss different themes relevant to their culture.

For diagnosis before the onset of the mental health disorders, Eichstaedt et al. \shortcite{eichstaedt2018facebook} use users' posts from Facebook to predict a future depression diagnosis. De Choudhury et al. \shortcite{de2013predicting} use a classifier to predict users' depression likelihood ahead of the onset of illness, with different measures used: language, linguistic style, emotion, ego-network, demographics and user engagement. Users suffering from depression tend to have less social activity, more negative emotions, more self-attention, concerns related to relations and medicine and more religious involvement.

We chose to tackle the problem of detecting early onset of depression from users' Reddit posts. To that end, we focus our efforts into processing the eRisk 2018 dataset \cite{losada2018overview}, given its success at the Workshop for Early Risk Detection on the Internet\footnote{\url{https://early.irlab.org/}} within The Conference and Labs of the Evaluation Forum (CLEF) and its fruitful submissions from participants. 

The teams from this workshop had different detection systems, based on bag of words ensembles \cite{trotzek2018word}, machine learning models with hand-crafted features \cite{trotzek2018word,ramiandrisoa2018irit,cacheda2018analysis,ramiirez2018upf} or with different text embeddings \cite{trotzek2018word,ramiandrisoa2018irit,ragheb2018temporal}, on sentence-level analysis to detect self references and extract different features \cite{ortega2018peimex}, on Latent Dirichlet Allocation (LDA) topic modelling \cite{maupome2018using}, models combining Term Frequency — Inverse Document Frequency with Convolutional Neural Networks \cite{wang2018neural} or other machine learning models. Most systems took the decision after the last chunk, only a few were able to emit a decision in the first chunks.

Several works addressing depression \cite{schwartz2014towards,resnik2015beyond} and PTSD \cite{coppersmith2015clpsych,preoctiuc2015role} use a topic modelling approach showing that topics encountered texts have important discriminative power to make the distinction between persons suffering from mental illnesses and healthy controls.

\section{Dataset}

Early Risk Detection on the Internet (eRisk) workshops organized by CLEF explore the technologies that can be used for people's health and safety and the issues related to building tests collections \cite{losada2018overview}. eRisk 2018 has two tasks, for early detection of depression and anorexia. We choose to focus on the task of detecting early onset of depression of social media users.

This task consists of sequentially processing chunks of Reddit posts from depressed users and controls. Submissions from each user are encoded in an xml file, one subject xml per chunk of data. Each xml contains the id of the subject and his posts and comments. Each submission has the posting time and the actual text. If a submission does not have a title, it is considered a comment. The goal is to detect depression as early as possible and the dataset has to be processed in chronological order. The test collection of posts from depressed and non-depressed users is split into 10 chunks. As training data, the teams had access to data from eRisk 2017, both train and test. The test chunks were released one every week. Every week the teams had to decide whether to label the user as depressed or non-depressed or to wait for the test data of the following week. 

The dataset contains 125 depressed users and 752 non-depressed users as training data and 79 depressed users and 741 non-depressed users as test data. The dataset has more posts and comments from people without depression than from users diagnosed with depression. From a total of 531,349 submissions, only 49,557 submissions are from users diagnosed with depression. The average time from the first to the last submission is between 2 and 3 years, so the posts were collected over a long period of time \cite{losada2018overview}.

\section{Method}

Our methodology for early diagnosis of depression follows a classical Natural Language Processing pipeline. To clean the users' texts, we transform them into lowercase, we remove the punctuation and stopwords, the numbers and URLs are replaced with specific tokens and we perform stemming with Porter Stemmer \cite{porter1980algorithm}. To reduce the dimension of the dictionary, we use collocations \cite{bouma2009normalized} to extract meaningful bigrams and trigrams.

The number of posts and comments from non-depressed users is much higher than those from depressed users. To balance the two classes, we downsample the majority class to a ratio of 2:1. 

We train our Latent Semantic Indexing model with 128 topics on every users' post. We use this model to extract topic modelling embeddings from users' texts and use them as input to our fully connected neural network architecture. The neural network has three hidden layers of 512, 256 and 256 neurons respectively, Leaky ReLU activation and we use Dropout for regularization. 
We use a random sample of 20\% of the training data provided by the organisers of the competition for validation.

The network has two outputs, one for classifying if the user is depressed or not and one for confidence estimation. The motivation for using this architecture is to learn the confidence \cite{devries2018learning} of our predictions and use it to make a decision on whether to label a user or wait for the next chunk of data. The learned confidence, besides its use case in out-of-distribution detection, can be used as a measure for how much the model trusts its classification output to be correct. As such, we consider the classification output only if the confidence exceeds a certain threshold. As indicated by DeVries et al. \shortcite{devries2018learning}, the network loss is computed by interpolating the predicted probabilities $p$ with the target $y$, using the computed confidence score $c$, as follows:

\begin{equation}
    p_i' = c \cdot p_i + (1 - c) y_i
    \label{eqn:p}
\end{equation}

The final loss is then given by:

\begin{equation}
    \Lb = - \sum_{i = 1}^{M} log(p_i')y_i - \lambda log(c)
    \label{eqn:loss}
\end{equation}

Where, in our case, $M = 2$, is the number of classes. The loss includes an additional term that forces the predicted confidence to be as high as possible. We performed an ablation study on the validation data on the confidence penalty $\lambda$.

A recent study by Hein et al. \shortcite{hein2019relu} shows that neural networks with ReLU activation functions tend to be overconfident on incorrectly classified samples, thus we can not rely only on the output probabilities, and the predicted confidence offers a more reliable measure of uncertainty of the classification.

As the number of submissions seen by the model increases, we want to make a decision as early as possible and thus we use a decaying function that decreases progressively the fixed threshold for confidence. The decision function is defined as follows:

\begin{equation}
    D_w(x) = \left\{
	\begin{array}{ll}
		\mbox{decide for}\, x  & \mbox{if } c > T * e^{-sw^2} \\
		\mbox{wait for data} & \mbox{otherwise }
	\end{array}
\right.
\end{equation}

Where $x$ is the embedding for the current users' posts, $w$ is the week number (i.e. the current chunk), $s$ is a scaling factor and $T$ is the initial threshold. We choose $T = 85\%$ and progressively scale it down to $40\%$. The scaling factor is computed such that, at the final chunk, the threshold is less than the smallest confidence encountered on the training data.

At the test phase, the proposed model does not make an independent decision for each chunk of data in the test set. In the first chunk of data, if the model is not confident enough to make a final decision regarding the depressed or non-depressed status of a user, then, starting with the second chunk of data, we concatenate the current chunk with the previously available chunks for the current user. This way, the LSI model has more data for making better informed predictions.

\section{Results}

Our results on eRisk 2018 dataset are presented in Table \ref{tab:results}. Even if $F_1$ is a standard evaluation measure used for imbalanced classification, it does not include the time component of the early detection task, thus Losada and Crestani \shortcite{losada2016test} propose an evaluation metric better suited for this task, the Early Risk Detection Error (ERDE). 

ERDE is defined as:
\begin{equation}
ERDE_{o}(d, k) = \left\{
    \begin{array}{llll}
        c_{fp} \quad  if \quad d=FP\\
        c_{fn} \quad if \quad d=FN \\
        lc_{o}(k) \cdot c_{tp} \quad if \quad d=TP\\
        0 \quad if \quad d=TN\
    \end{array}
\right.
\label{eqn:erde}
\end{equation}

The use of false positive (FP), false negative (FN), true positive (TP) and true negative (TN) for prediction \textit{d} is to avoid the classifiers that always predict the label of the majority class.  $ lc_{o}(k)\in [0, 1] $ encodes a cost for the delay in detecting TP. For the eRisk datasets, where the number of negative labels is greater than positive labels, the value of $ c_{fn} $ is 1 and $ c_{fp}$ is 0.1296, set according to the proportion of depressed users in eRisk 2017 dataset \cite{losada2018overview}.  $ c_{tp} $ is set to  $ c_{fn} $ because the late detection of people at risk of depression can have serious consequences, a late detection is considered as equivalent to not detecting the depressed user at all. The late detection of TN cases does not affect the effectiveness of the system. 

The goal of the system is to detect as early as possible people at risk of depression. For the detection of non-depressed users, the time of the detection is not relevant.
The latency cost function, which grows with \textit{k} (the number of submissions seen by the algorithm), is defined as:

\begin{equation}
    lc_{o}(k)= 1 - \frac{1}{1 + e^{k - o}}
    \label{eqn:lc}
\end{equation}

\textit{o} represents the number of posts after which the cost grows more quickly.

\begin{table}[hbt!]
    \centering
    \small
    \renewcommand{\arraystretch}{1.1}
    \begin{tabular}{p{0.25\linewidth}|p{0.11\linewidth}p{0.13\linewidth}|p{0.06\linewidth}p{0.06\linewidth}p{0.06\linewidth}}
     \textbf{Method} & \textbf{ERDE$_5$} & \textbf{ERDE$_{50}$} & \textbf{F$_1$} & \textbf{Prec} & \textbf{Rec}\\
    \hline \hline
    Baseline LSI  & \textbf{9.98\%} & 8.29\% & 0.25 & 0.22 & 0.29 \\
    \hline \hline
    $\mbox{LSI}_c$ $\lambda = 0.01 $  & 14.19\% & 11.41\% & 0.25 & 0.15 & 0.87 \\
    \hline
    $\mbox{LSI}_c$ $\lambda = 0.1 $  & 11.12\% & 9.09\% & 0.28 & 0.20 & 0.48 \\
    \hline
    \textbf{$\mbox{LSI}_c$} $\mathbf{\lambda = 0.2} $  & 10.24\% & \textbf{7.74\%} & \textbf{0.30} &\textbf{0.25} & 0.38 \\
    \hline
    $\mbox{LSI}_c$ $\lambda = 0.4 $  & 11.15\% & 8.53\% & 0.25 & 0.17 & 0.47 \\
    \hline
    $\mbox{LSI}_c$ $\lambda = 0.6 $  & 12.67\% & 10.17\% & 0.25 & 0.15 & \textbf{0.71} \\
    \hline
    $\mbox{LSI}_c$ $\lambda = 0.8 $  & 10.53\% & 8.08\% & 0.30 & 0.21 & 0.56 \\
    \hline \hline
    Funez et al.\shortcite{funez2018unsl} & 8.78\% & 7.39\% & 0.38 & 0.48 & 0.32 \\
    \hline
    Trotzek et al.\shortcite{trotzek2018word} & 9.50\% & 6.44\% & 0.64 & 0.64 & 0.65 \\

    \end{tabular}
\caption{Classification results on the detection of early onset of depression task from eRisk 2018 dataset.}
\label{tab:results}
\end{table}

The detection task is difficult, as seen in the low values of $F_1$ and Precision. However, the task is to predict \textit{early} onset of depression, and for that, the ERDE metrics are more appropriate, as they are a measure of prediction delay. ERDE$_5$ metric is very sensitive to delays, after the first 5 submissions from the user the penalties grow quickly. In contrast to ERDE$_5$, for ERDE$_{50}$ the penalties grow only after the first 50 submissions from the user. The difference between ERDE$_5$ and ERDE$_{50}$ is very important in practice because of the consequences of late detection of depression signs. As the task suggests, the detection should be made as early as possible.

To measure the impact of our learned out-of-distribution confidence from the neural network, we also trained a plain ReLU network with cross-entropy loss. For this model, we employed a hard threshold on the output probabilities for whether to wait for more data or classify the sample. As shown by Hein  et  al. \shortcite{hein2019relu}, ReLU networks can be overly confident on misclassified examples. This is shown in Table \ref{tab:results}: the model has a low $ERDE_5$ score as the output probabilities mostly have extreme values, which means that for most users the model makes a decision from the first chunk of data. 

We trained our model with different $\lambda$ values in order to see the impact of the confidence component on the results. Larger values for $\lambda$ make the model overly confident, as expected from Equation \ref{eqn:loss}, the best performing model being the one with $\lambda = 0.2$. Smaller values of $\lambda$ generate a wider confidence distribution on the training examples, facilitating the decision process, as extreme values either make the model overly-confident on every example, or not confident at all. This is consistent with findings by DeVries et al. \shortcite{devries2018learning}.

In Table \ref{tab:results} we also present the best two submission from the eRisk 2018 Workshop, the one from Funez et al. \shortcite{funez2018unsl}, having the best results for the ERDE$_5$ metric, and the one from Trotzek et al. \shortcite{trotzek2018word} having the top ERDE$_{50}$ score.

We can assume from these results that topics encountered in user writings have important discriminatory power. Depressed users mostly write about different subjects than non-depressed subjects, consistent with results from the work of Resnik et al. \shortcite{resnik2015beyond}. The writings from users diagnosed with depression are more focused on their feelings and their life events. Topics related to those themes contain words such as \textit{someone kill, bad though, never able to get, forever alone, life save, stay sober, i am sad, still can't, improve life. new hope, oneself, tell anything, happy sad, hope one day}. Texts from non-depressed users are found in topics related to their hobbies containing specific words: \textit{black mirror, first season, movie adaptation, hologram, nine inch nails, jimi hendrix, artist name, vlog, game, fallout, terra mistica, way to make money, paid time, really proud, amazon whishlist, food industry, white bread}. 

\section{Conclusion}

In this paper, we use the eRisk 2018 dataset on Early Detection of Signs of Depression for depression classification from Reddit posts. Our method uses Latent Semantic Indexing for topic modelling and to generate the embeddings used as input for our neural network, but focuses on using a learned out-of-distribution confidence score alongside the classification output to decide whether to label the user or wait for more data. Besides its initial use case in out-of-distribution detection, we repurposed the confidence score as a measure for how much the model trusts its classification output to be correct. We showed that, in general, there is a significant difference in writing topics depending on the users' mental health, to the extent that it contains enough information for use in classification.

\section*{Acknowledgements}

We would like to thank our reviewers for their useful comments and suggestions that helped us improve this paper and also to the organizers of the eRisk Workshop for their efforts in encouraging the research on mental illnesses detection from social media.

\bibliographystyle{acl}
\bibliography{refs}

\end{document}